\documentclass{article}

\usepackage{arxiv}

\usepackage[utf8]{inputenc} 
\usepackage[T1]{fontenc}    
\usepackage{hyperref}       
\usepackage{url}            
\usepackage{booktabs}       
\usepackage{amsfonts}       
\usepackage{nicefrac}       
\usepackage{microtype}      
\usepackage{lipsum}
\usepackage{graphicx}
\graphicspath{{media/}}     

\usepackage{natbib}
\usepackage{doi}
\usepackage{adjustbox}
\usepackage{amsmath}
\usepackage{amssymb}
\usepackage{float}
\usepackage[T1]{fontenc}
\usepackage[utf8]{inputenc}
\usepackage{mathtools}
\usepackage{multicol}
\usepackage{multirow}
\usepackage{wasysym}

\title{Free will belief as a consequence of model-based reinforcement learning}

\date{November 14, 2021}	

\author{
    Erik M.  Rehn\\
    Independent Researcher\\
    Malmö, Sweden. \\
    \texttt{erik.m.rehn@gmail.com} \\
}

\hypersetup{
pdftitle={Free will belief as a consequence of model-based reinforcement learning},
pdfsubject={q-bio.NC, q-bio.QM}, 
pdfauthor={Erik M.  Rehn},
pdfkeywords={free will belief, reinforcement learning, model-based, credit assignment},
}

\begin{document}
\maketitle

\begin{abstract}
The debate on whether or not humans have free will has been raging for centuries. Although there are good arguments based on our current understanding of the laws of nature for the view that it is not possible for humans to have free will, most people believe they do. This discrepancy begs for an explanation. If we accept that we do not have free will, we are faced with two problems: (1) while freedom is a very commonly used concept that everyone intuitively understands, what are we actually referring to when we say that an action or choice is ``free" or not? And, (2) why is the belief in free will so common? Where does this belief come from, and what is its purpose, if any? In this paper, we examine these questions from the perspective of reinforcement learning (RL). RL is a framework originally developed for training artificial intelligence agents. However, it can also be used as a computational model of human decision making and learning, and by doing so, we propose that the first problem can be answered by observing that people’s common sense understanding of freedom is closely related to the information entropy of an RL agent's normalized action values, while the second can be explained by the necessity for agents to model themselves as if they could have taken decisions other than those they actually took, when dealing with the temporal credit assignment problem. Put simply, we suggest that by applying the RL framework as a model for human learning it becomes evident that in order for us to learn efficiently and be intelligent we need to view ourselves as if we have free will.

\end{abstract}

\keywords{ free will belief \and reinforcement learning \and model-based \and credit assignment}

\section{Introduction}

Although, according to the current understanding of the laws of nature, there are good arguments in support of the view that it is not possible for humans to have free will, so-called free will belief is widely spread across cultures (Wisniewski et al., 2019). This discrepancy between the materialist view held by large parts of the scientific community, on the one hand, and the subjective experience most laypeople have of their own freedom, on the other hand, begs for an explanation. If we assume that the materialists are right (i.e., that we lack free will), we must ask ourselves the following questions: 

\begin{enumerate}
	\item What does ``freedom" mean when used in everyday language? Freedom is a commonly used concept but what are we actually referring to when we say that an action or choice is ``free" or not? 

	\item Why is free will belief so common? Even the most determined determinist feels as if he or she can make free choices. Where does this experience come from, and what is its purpose, if any?
\end{enumerate}
In this paper, we attempt to address these questions from the perspective of an agent trained with model-based reinforcement learning (RL) (Silver et al., 2017; Levine and Koltun, 2013; Deisenroth and Rasmussen, 2011). Reinforcement learning is a framework for training artificial intelligence (AI) agents in settings where there is a delay between the actions of the agent and the evaluation of the objective of the agent. The purpose of an RL agent is to learn to perform sequences of actions that maximize the long-term reward it receives from its environment. Where reward is received when the objective assigned to the agent is fulfilled. A classic example is games like chess, where an AI should learn to make a sequence of turns that lead to a win.

Importantly, reward maximization through reinforcement learning has been proposed not only as a method for building AI systems, but also as a computational theory of human learning and decision-making (Silver et al., 2021). Although the human brain most definitely mixes many different kinds of algorithms and objectives, reward maximization is suggested as a general framework for how to express the multitude of objectives. The pursuit of reward allows humans to optimize for long-term success, which is subjectively defined through the reward system of the brain—a system that has evolved to balance different needs and desires and preserve the homeostasis of the body. Accordingly, reward is hypothesized to be a ``common currency" that allows for most (if not all) aspects of intelligence and behavior to be prioritized and integrated.

In the literature, it has frequently been argued that the problem of free will is a semantic issue. What do we actually mean with \textit{free} will? (e.g., see the debate between Dennett (2017) and Harris (2012)). In this paper, we agree that this is indeed the case. The assumption that humans are reward-maximizing RL agents offers a possibility to resolve the conceptual confusion. We argue that, whenever we talk about freedom, we are conflating two different concepts. The first of these concepts—which we will refer to here as \textit{physical freedom}—is what incompatibilists have in mind when they conclude that the currently known laws of nature do not permit one to have free will in any meaningful way. This kind of freedom is essentially randomness. It is the possibility for things to happen without a cause, which is in physics a consequence of quantum mechanics, or at least according to some interpretations thereof. Therefore, it makes perfect sense to conclude, as hard incompatibilists do, that physical freedom is not something we can possess, since if something is random, it is by definition not under our control (Pereboom, 2001).

However, as we will show, the RL perspective allows us to talk about and formally define the freedom of an agent in a different yet meaningful way. We refer to this other type of freedom as \textit{value freedom}. In this term, ``\textit{value"} does not refer to values as in \textit{moral values}, but rather to the estimated values of actions available to an RL agent given its current knowledge of the environment. The value freedom of an agent is defined as the information entropy of its action selection process and thus can, at least in some sense, be viewed as a measure of unpredictability, regardless of whether the selection process is fully deterministic or somewhat stochastic. Accordingly, we suggest that the concept of \textit{value freedom} bridges the everyday use of the word \textit{free}, the compatibilist view that it is meaningful to talk about free will even under strict determinism, and the incompatibilist view according to which our understanding of causality does not permit us to have free will in the physical sense. 

Furthermore, we also suggest that value freedom is beneficial for agents due to its connection to the exploitation vs. exploration dilemma faced by all agents trying to learn from delayed rewards (Thompson, 1933; Lai and Robbins, 1985). High freedom states are desirable because they signal low-risk exploration and provide room for potential policy improvements.

Next, we argue that the RL perspective sheds light on why humans so strongly believe they have free will and that they can make decisions independently of causes outside of themselves. Central to efficient RL is a solution to the so-called temporal credit assignment problem (Minsky, 1963). Since rewards are in general sparse and delayed, an RL agent has to figure out which of all the actions leading up to the reward were actually important for the outcome. However, at present, most commonly used RL algorithms rely only on temporal proximity heuristics where the actions closer in time to the reward are simply assigned more credit than those farther away. This trick obviously breaks down in more complex environments with long temporal dependencies, as well as when different tasks are interleaved with each other. In such complex settings, in order to reinforce the correct behaviors, agents must figure out which actions were actually important, and why the reward was received. Therefore, credit assignment becomes a problem of counterfactual reasoning using a causal world model. Agents need to answer questions like ``If I did A instead of B, would I still receive reward C?" In other words, agents must be able to simulate or imagine having made choices they did not in fact make. We suggest that this necessary—albeit, from a deterministic point of view, flawed—self-model can explain the human experience of being free. We tend to think of ourselves as being capable of breaking the causal structure of our environment and making free choices, and this imagined freedom—despite being, in all probability, just a figment of imagination—is necessary for efficient learning and thus a vital part of our intelligence.

In what follows, we start with a description of model-based RL and the challenges that learning agents, including humans, face when trying to maximize reward. Then, based on the formulated definition of RL, we proceed to define the value freedom of an agent and, through a series of examples, demonstrate that this quantity closely relates to the concept of freedom, as used in everyday language. In the next section, we discuss why the credit assignment problem makes it necessary for efficient model-based RL agents to believe that—or at least act as if—they have free will that defies causality.

\section{Reinforcement learning}

To define the value freedom of an agent, we start by giving a formulation of RL. In RL, an agent is imagined to exist in an environment with a high-dimensional and time-evolving state, \( \mathbf{s}_t \). However, in the general case, this state is only partially observed by the agent with observations denoted by \( \mathbf{o}_t \). The agent also has a number of potential actions, \( \mathbf{a}_t \)\textit{,} it can choose from at each time step. These actions alter the observed state by changing both the environment and the part of the environment that is observed, hence \( \mathbf{o}_{t+1} = f(\mathbf{s}_t, a_t) \). Furthermore, the environment emits a scalar reward/penalty to the agent at each step, \( \mathbf{r}_t \in \mathbb{R}^1 \), with \( \mathbf{r}_{t+1} = f(\mathbf{s}_t, a_t) \). An RL agent’s objective is to find an action selection rule, or a so-called policy, \( a_t=\pi(\mathbf{o}_t) \), such that it maximizes the expected long-term reward the agent receives:

\begin{equation}
\label{eq:value}
V^*(\mathbf{s}) = \max_{\pi}\operatorname E[r | \mathbf{s}, a, \mathbf{\pi} ] \in \mathbb{R}^1
\end{equation}

where \( V^*(\mathbf{s}) \) is the expected reward or \textit{value} received by the agent when starting in state, \( \mathbf{s} \)\textit{,} and choosing a sequence of actions, \( a_{1...t} \), according to the optimal policy \( \pi^* \). Hence, the goal of RL is to find a policy, \( \pi \), which as close as possible result in the same expected reward as \( \pi^* \).

To break down the problem further, we also define the \textit{action value} of an action in state \textbf{\textit{s}}, as follows:

\begin{equation}
\label{eq:nolabel_1}
Q^*(\mathbf{s},a) = \operatorname E[r | \mathbf{s}, a, \mathbf{\pi}^* ] \in \mathbb{R}^1
\end{equation}

which is the expected reward received over time if an action, \( a \), is chosen according to \( \pi^* \), starting in state \( \mathbf{s} \). Hence, one way to maximize \( V^*(\mathbf{s}) \) is to try to estimate \( Q^*(\mathbf{o}, a)\) and then choose the actions with the highest value at each time step. If \( Q^\pi = Q^\pi(\mathbf{o}, a)\) is the agent’s estimated action values given its currently observed state \( \mathbf{o} \), and its policy is \( a=\pi(\mathbf{o}) = \text{argmax}(Q^\pi(\mathbf{o}, a)) \), RL comes down to learning to estimate \( Q^\pi \)\textit{ }as closely as possible to the optimal action values, \( Q^* \).

As a side note, observe that the reward signal, \( r_t \),\textsubscript{ }can be viewed as a special dimension of \( \mathbf{s}_t \) that attains its status from the objective of RL to maximize the expected reward. In the case of humans, the reward system of the brain can thus be viewed as a special part of the environment whose output the learning system of the brain tries to maximize.

The challenge of RL arises in environments where the reward signal is sparse, i.e., mostly zero, which is generally the case. For instance, in a game of chess, you do not receive a reward until you have performed a long sequence of actions that result in a win or loss. Owing to this delay between actions and rewards, RL presents a learning agent with two central challenges: reward prediction and credit assignment.

If an agent can predict how its behavior will alter its environment and what future states will result in reward, it can use these predictions to better choose its actions, i.e., estimate \( Q^\pi \) as accurately as possible. Less trial-and-error is needed if an agent, when faced with a new situation, is capable of seeing into the future to make predictions of the outcome of its actions.

In contrast to planning that requires an RL agent to predict the future, the problem of credit assignment requires it to understand the past. If a sequence of actions leads to a successful outcome, the agent needs to figure out which actions were actually important for the success, and which were not. Accordingly, the agent has to assess which behaviors shall be reinforced, making them more likely to be repeated in similar situations in the future, and which shall be suppressed.

Model-based RL (Silver et al., 2017; Levine and Koltun, 2013; Deisenroth and Rasmussen, 2011) is an approach to tackle both the problem of reward prediction through planning and/or that of credit assignment by equipping the agent with some form of generative model of its environment, i.e., a world model. Specifically, reward prediction can be achieved by running the world model forward, allowing the agent to search for favorable action sequences without really performing them, while credit assignment is attained by employing the world model to evaluate which past actions were actually important for the received reward, without needing to go back and test alternative sequences. Accordingly, credit assignment can be viewed as a problem of causal inference (Pearl, 2010), where actions causally necessary for reaching the rewarding state are identified and assigned credit. Both prospection and retrospection through a causal world model are thus crucial for effective RL.

\section{What is freedom?}

Similar to many other concepts used in everyday language, \textit{freedom} and \textit{voluntary} vs. \textit{involuntary} action are not precisely defined. However, this does not make them less useful. For instance, the United Nations (UN) Universal Declaration of Human Rights (UN, 1948) is sprinkled with the words \textit{free} and \textit{freedom}. 

Yet, if we seriously consider the perspective that humans are RL agents, a problem emerges. Namely, what does it mean for an RL agent to make free choices? An RL agent is just an algorithm or a collection thereof, be it deterministic or somewhat stochastic, or equipped with a world model or not. An RL agent is still just a mapping between states of the environment and actions. It is difficult to talk about algorithms having free will.

We suggest that the solution to this problem starts by noting that the action value estimates, \( Q^\pi(\mathbf{o}, a)\)\textit{,} represent something close to the ``will" of an agent, since in every situation an agent selects the action with the highest expected value. Therefore, action values represent what an agent wants to do, i.e., what outcomes it expects to be rewarding or least costly, and how effective it believes the available actions are when trying to reach those outcomes. For instance, if the agent is an imprisoned human, he or she might really want to escape, but, at the same time, realizes that the probability for success is minuscule and thus estimates the action value of escaping to be rather small. Therefore, after an agent has considered the probability of success, the resulting action values can be viewed as the will of that agent.

Based on the aforementioned definition of ``will", we suggest that it is also possible to quantify how free an agent is. To this end, we define the action selection probability as:

\textit{  \begin{equation} \label{eq:prob_a} P(a_i) = \text{softmax}(Q^\pi(a_i)) = \frac{\exp(Q^\pi(a_i))}{\sum_j \exp(Q^\pi(a_j))} \end{equation} }

where the softmax function normalizes\footnote{The exact method of normalization in not important. We here ignore the problem that the softmax function is not scale invariant.} the estimated action values into a distribution over actions. We then use this to define the freedom of an agent with an observed state as the information entropy of the action selection distribution:

 \begin{equation} \label{eq:entropy} H_a = - \sum_iP(a_i) log_2(P(a_i)). \end{equation}

This quantity, which we refer to as \textit{value freedom}, can thus be interpreted as the uncertainty of an agent’s action selection if the policy samples action weighted according to \textit{P(a)}. It is high in situations where the agent’s estimated action values, \( Q^\pi(\mathbf{o}, a)\) , are uniform, and low when the value of one or a few actions appears to be markedly higher than those of all other actions. 

However, note that for this treatment of RL it does not matter whether the policy is stochastic or deterministic, as long as we can define the relative values of actions given a state, \( Q^\pi(\mathbf{o}, a)\), and normalize these into an action selection distribution, \( P(a) \). For instance, the policy, \( \pi(\mathbf{o}) \), can be greedy as above and simply select the most valuable action, or stochastic and randomly sample actions according to \textit{P(a)} somehow. Stochasticity can also enter through noise in the estimation of \( Q^\pi(\mathbf{o}, a)\)\textit{ }rather than in the action selection process.

To demonstrate that value freedom, \( H_a \) \textit{, }is indeed a quantity that closely maps to what, in everyday language, is referred to as freedom of choice, let us imagine two idealized scenarios where one could consider a binary choice to be free or not (here we switch to a second-person perspective to allow the reader to better imagine themselves in these situations):

\begin{enumerate}
	\item You are sitting on a lush green lawn. Someone asks you in a friendly way to raise your left arm if you like, with no consequences whether you do it or not.

	\item You are standing at the edge of a precipice with a deep dark ocean crashing against the sharp cliffs beneath you. Someone sneaks up behind you and threatens to push you off the ledge if you do not raise your left arm.

\end{enumerate}
Let us also assume you can be very certain that what is promised will also happen, and that you are not suicidal.

When comparing these scenarios, one could argue that you are not free to choose what to do in either case, because you are just an algorithm and, as such, have no free will. Whether you raise your arm or not is just a result of the initial conditions of the situation and some deterministic or stochastic machinery in your head. However, in the common-sense meaning of being free, it is obvious that you are freer to choose what to do in the first scenario. Nothing is forcing you to do anything and, if you raise your arm, it seems as if you were free to not do it. So what is the difference between these two situations?

The central argument here is that the reason behind our perception of these two cases as very different is not that you are actually more physically free in one case or the other, but rather that you estimate the action values, \( Q^\pi(\mathbf{o}, a))\)\textit{, }differently. Your goal (or that of any other RL agent) is to maximize the reward you receive over time. To this end, you need to continuously estimate the future reward of every available action and select the action with the highest estimate. Hence, you always perform the action that you subjectively estimate to be the best, i.e., the one that, in your assessment, has the highest estimated action value.

From this perspective, the difference between the two scenarios lies in how you reasonably would estimate the future reward of raising your arm or not. In the first scenario, raising your arm has no consequences; therefore, doing so or not has basically equal values (if we disregard that the tiny bit of physical exercise probably would do you good...). In contrast, in the second scenario, your fear of death produces a very large estimated negative reward for the option of not raising your arm. 

For more examples of how various value estimates, in binary choice situations, result in different levels of value freedom, see Figure 1 and Table 1.

 Note that you could also imagine that your policy is random, and that you do not always select the action with the highest value, but instead somehow sample your actions weighted according to the estimated rewards, since your action selection process might be more or less noisy. Nevertheless, intuitively, the ``freedom" of your policy is large in the first scenario and very small in the second scenario, because, in the former case, there is simply less difference in value between available options. Indeed, there is empirical evidence showing that a person indifferent to the consequences of his/her choice is seen by lay people as maximally free (Deutschländer, 2017).

\begin{center}
\includegraphics[width=7.43cm,height=13.06cm]{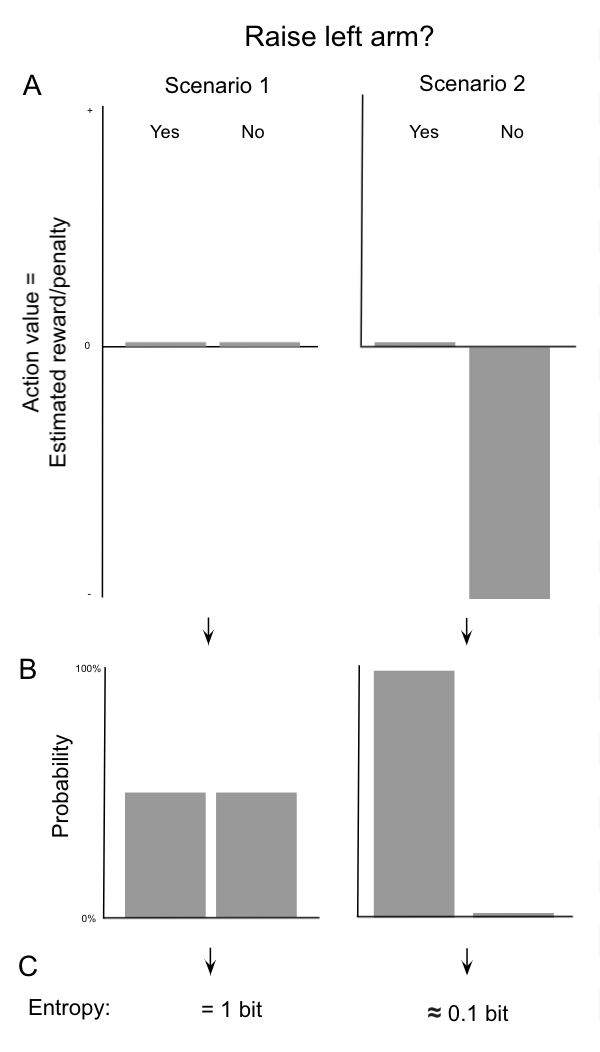} \\ Figure 1: (A) How most people would reasonably estimate the reward and penalty, i.e., the value, for the option of raising their left arm as requested in scenarios 1 and 2. In scenario 1, both alternatives would lead to a zero reward. In scenario 2, not raising the arm would likely lead to a very large penalty (death). The exact scale of the reward is not important here. (B) The action values converted into probability distributions. These can be viewed as probabilities for raising the arm or not, although the actual action selection is deterministic. (C) The Shannon entropy for the probability distributions in B. For a uniform distribution with only two options, the maximum entropy is 1 bit. If there is a 99\% chance of raising the arm in scenario 2, the entropy is approximately 0.1 bit.
\end{center}

\begin{table}[!htbp]
\renewcommand{\arraystretch}{1.3}
\begin{adjustbox}{max width=\textwidth}
\begin{tabular}{p{2.88cm}p{2.75cm}p{3.41cm}p{6.85cm}p{2.88cm}p{2.75cm}p{3.41cm}p{6.85cm}}
\hline
\multicolumn{1}{|p{2.88cm}}{Value of 1st choice, \( Q^\pi(\mathbf{o}, a_1)\)} & 
\multicolumn{1}{|p{2.75cm}}{Value of 2nd choice, \( Q^\pi(\mathbf{o}, a_2)\)} & 
\multicolumn{1}{|p{3.41cm}}{Value freedom, \( H_a \)} & 
\multicolumn{1}{|p{6.85cm}|}{Interpretation} \\ 
\hline
\multicolumn{1}{|p{2.88cm}}{0} & 
\multicolumn{1}{|p{2.75cm}}{0} & 
\multicolumn{1}{|p{3.41cm}}{High} & 
\multicolumn{1}{|p{6.85cm}|}{You are indifferent to \( a_1 \) or \( a_2 \)} \\ 
\hline
\multicolumn{1}{|p{2.88cm}}{0} & 
\multicolumn{1}{|p{2.75cm}}{+} & 
\multicolumn{1}{|p{3.41cm}}{Low} & 
\multicolumn{1}{|p{6.85cm}|}{a\textsubscript{2 } seems to result in a reward you cannot afford to not choose (e.g., money, or pleasure).} \\ 
\hline
\multicolumn{1}{|p{2.88cm}}{0} & 
\multicolumn{1}{|p{2.75cm}}{-} & 
\multicolumn{1}{|p{3.41cm}}{Low} & 
\multicolumn{1}{|p{6.85cm}|}{a\textsubscript{2} seems to result in a penalty (e.g., physical violence (pain)).} \\ 
\hline
\multicolumn{1}{|p{2.88cm}}{+} & 
\multicolumn{1}{|p{2.75cm}}{+} & 
\multicolumn{1}{|p{3.41cm}}{High} & 
\multicolumn{1}{|p{6.85cm}|}{All choices seem good.} \\ 
\hline
\multicolumn{1}{|p{2.88cm}}{-} & 
\multicolumn{1}{|p{2.75cm}}{-} & 
\multicolumn{1}{|p{3.41cm}}{High} & 
\multicolumn{1}{|p{6.85cm}|}{Bad state to be in, but you are still free to choose between plague or cholera. We often view these states as unfree, since we imagine that there is a third neutral option (no plague and no cholera). } \\ 
\hline
\multicolumn{1}{|p{2.88cm}}{+} & 
\multicolumn{1}{|p{2.75cm}}{-} & 
\multicolumn{1}{|p{3.41cm}}{Very low} & 
\multicolumn{1}{|p{6.85cm}|}{\( a_1 \)is undoubtedly and massively better than \( a_2 \). \( a_2 \) is probably not even considered as an option.} \\ 
\hline
\end{tabular}
\end{adjustbox}
\caption{Table 1: How possible estimated action values result in different levels of value freedom in a binary choice situation.  ``+" and ``-" refer to large positive or large negative action values. }
\label{tab:table_1_how_possible_estimated}\end{table}

At this point, one might rightfully object that, thus far, this seems like a very narrow definition of freedom. Isn’t it possible to be very sure about what you want to do, but still feel free to do so or not? When we talk about freedom, are we really referring to uncertainty in our action selection? For instance, let us imagine the following scenario:

You sit down in a restaurant and are handed the menu. You can pick anything from it. You are free to choose your dinner: you have no allergies, and you do not have any particular preferences this evening. After casually skimming through the menu at random, you decide to have meatballs with mashed potatoes. You sip on your drink and think about something else. When several minutes later the waiter comes over to take the order, he does not accept your choice right away but instead suggests you should try the fried herring. It is supposedly much better. Immediately, you feel the annoyance grow inside. Who would ever want fish when they could have meatballs?!

In the above scenario, you could argue that you feel completely free to make whatever decision you want, but you still prefer one particular option over the others; hence, what we previously defined as value freedom cannot be what our understanding of freedom is based on. However, we would like to argue that the reason for this apparent contradiction is two-fold.

First, you feel free to choose whatever you want to have for dinner because it really does not matter much. Even if you are disappointed when you cannot have meatballs, you know that the relative reward between having herring, meatballs, or any other well-cooked dish is marginal. In contrast, in a scenario where, after starving for a week, you have the choice between having meatballs and no dinner at all, your freedom is quite limited, to say the very least. Hence, the value freedom is actually relatively high in the original scenario.

Second, the value freedom was indeed very high when you first sat down and started skimming through the menu. Since you had no initial preferences, your memory of the decision is that you were free to pick anything. However, once taken, your decision somehow changed your preferences, and suddenly meatballs seemed much more rewarding than what else was on the menu. Accordingly, when the waiter tried to change your mind, your value freedom decreased, since you were now less probable to choose the herring or anything else. In this way, our sense of being free to make decisions is based not only on our current value freedom, but also on our memory of the causal chain of decisions we made, and the value freedom in those situations. Note that, for an agent without any memory of past choices, this would not be the case.

The purpose of the examples provided above is to show that our common-sense understanding of freedom comes from how we (i.e., our reward system and world model) subjectively estimate the values of available actions, rather than from how free we are in a physical sense. It is this type of freedom that we refer to in everyday language when we say that someone is free to choose or forced to do something. If someone is threatened with violence, they tend to do what they are told and, therefore, have low value freedom since their actions become predictable.

However, there is one additional aspect that needs to be accounted for when we assess if someone’s actions were free or not. First, note that the value freedom can be defined for any system— be it a chess computer or an insect—that evaluates different options and somehow selects between them. Furthermore, the value freedom per se does not depend on the quality of the evaluation. A weak chess computer that plays on a beginner’s level can on average have the same value freedom as the one playing like a grandmaster. Similarly, the intoxicated you can have the same value freedom as the sober you.

Therefore, when judging the freedom that we or someone else possess when faced with a decision, we do so taking into account not only the relative estimated values of available options but also the quality of the estimates. Accordingly, our common-sense understanding of when a person makes a voluntary action according to his or her free will relies on both the assumption that the value freedom is high \textit{and} that the person’s competence to judge the relative value of available options is roughly equal to what can be expected from an adult human being in a similar situation and culture. Using our world models and counterfactual reasoning, we try to understand how the person would have predicted his or her future rewards, given that the person, for example, was not intoxicated or suffering from any other condition that could have reduced the quality of his or her predictive power. Other common reasons for us to not judge someone’s actions as free are deception and young age. Similarly, we analyze our memory of our own actions and their causes and might feel that we did not do something out of free will, since we misjudged the situation due to some external factors beyond our control.

Another exception from only considering the momentary value freedom is how the future outcome of actions influence our perception of freedom. A consequence of quantifying freedom as entropy is that splitting the action space into more actions with the same or very similar outcome increases the freedom. An agent faced with choosing between two  actions that result in the same state is obviously not free. In such situations one must consider not only the value of actions but also the diversity of their outcomes. However, this is subjective to the agent. Two agents can have completely different opinions on whether outcomes are similar or not depending on their reward function. And, further note that outcome is not equivalent to received reward. Actions can have the same estimated value but still result in very different future states. For instance, imagine choosing between two vacation destinations. You might estimate both as equally likely to give you pleasant experiences, but you still know that when you come home these experiences will have been vastly different.

In summary, we argue that the freedom of an agent can be quantified by looking at the uniformity of \( Q^\pi \). The amplitudes of \( Q^\pi \) naturally translate to the will of the agent, since they represent how beneficial each action appears to be to the agent, while the uniformity of \( Q^\pi \), here suggested to be measured through the information entropy, \textit{H\textsubscript{a}}, corresponds to how ``free" the will is. In other words, value freedom could also be viewed as the ``entropy of the will". And as such, its unit, bits, corresponds to how much information is gained by observing the choice of an agent. Since the decisions where the value freedom is high involve more uncertainty concerning what the agent will do, the information content is higher than for low freedom decisions.

In addition to the momentary value freedom, we also suggest that our common-sense understanding of whether an agent's action given a state is free is also based on the following factors:

\begin{enumerate}
	\item The historic value freedom: How high was the value freedom in situations leading up to the current state? A decision is rarely made in isolation.

	\item The quality of the action values: Does the agent estimate action values, and thus the value freedom, as well as one can expect?
	
	\item The diversity of the outcomes: Does the agent consider the outcomes of available actions different enough for the actions to be seen as separate choices?

\end{enumerate}
Note that all these factors are highly subjective, from the perspective of both the agent itself and other agents judging someone else’s freedom. Value freedom is inherently subjective, while physical freedom is objective.

\section{Why do we believe we are free?}

Now that we have a way to quantify and think about our subjective understanding of freedom through the concept of value freedom, we will address the other questions we set out to answer: Why do we feel that we have free will, and what is the purpose of this feeling? Two questions which we suggest can be translated to: (1) Why do we tend to mix up value freedom with physical freedom? And, (2) why is high value freedom desirable?

We hypothesize that the answer to (2) can be found in the need for RL agents to deal with the famous exploration vs. exploitation dilemma (Thompson, 1933; Lai and Robbins, 1985). High value freedom indicates to an agent that the risk of exploration is low, because, in such situations, it looks as if there is not much to lose from exploring. Since several alternatives have similar predicted values, choosing one that is not known to be optimal is perceived as less of a risk. From this perspective, freedom can be seen as room for learning. When value freedom is high, an agent has the opportunity to test new approaches that hopefully will improve its world model, leading to an improved policy. Hence, high value freedom signals potential low-risk policy improvements.

To exemplify, let us reuse the scenario where you sit down at a restaurant and are about to choose what to have for dinner. If you do not have any particular preferences this evening, you are free to try something new, which may turn out to be your new favorite dish. However, if you are on a strict diet that limits your options, the likelihood of you discovering something new is obviously much lower. Hence, less value freedom presents fewer opportunities for learning.

But, note that the value freedom does not say anything about how good your options are in an absolute sense—rather, it only shows how you subjectively value them relative to each other. Therefore, if all dishes on the menu look equally disgusting to your refined taste, then you are still bad off, even if your freedom is high. The inductive bias of ``freedom desire" is thus balanced towards other goals, like having good food, through the common currency of reward. We are all familiar with the fact that people are often willing to give up some of their freedom in exchange for the fulfillment of other more basic needs.

We hypothesize that, rather than directly giving an agent a higher reward, striving for freedom makes the policy more robust and increases the venues for improving it in the future. Similarly to curiosity (the intrinsic reward we get when we learn something new, i.e., improve our world model) (Schmidhuber, 1991; Deepak et al., 2017), our sense of freedom can thus be viewed as an inductive bias that evolution has given us to increase our learning efficiency. On this view, freedom is also intimately related to curiosity. Only a free learning agent can afford to be curious.

Intuitively, this makes sense. If you are always forced to do the same thing over and over again, there is no room to try anything new, and you will never learn. However, an important distinction to be made here is that the lack of freedom does not necessarily mean the lack of exploration. We all have been in situations where our exploration was guided by a teacher, and where we trusted our teacher to show us the best path forward. When we follow a course or a curriculum, this is what happens. We suggest that \textit{supervised exploration} can be seen as a middle-ground between imitation learning and reinforcement learning, where the teacher does not explicitly demonstrate what the correct answer or solution is, but instead points the learner in the right direction by supplying problems of suitable difficulty. In this mode of learning, although the learner’s value freedom can be low, exploration and learning still happen, and such learning can often be more efficient than when the learner relies only on unsupervised exploration. This is why schools were invented. During supervised exploration, the teacher employs its model of the learning domain and the learner and, through authority or trust, skews the learner’s reward estimates to make some actions look more preferable than others.

Now that we have discussed a potential motivation for the importance of (value) freedom to learning agents like humans, let us turn to the other question: Why do we tend to mix up value freedom with physical freedom?

The short answer is: Because we constantly imagine that we have physical freedom. This fantasy, although arguably rather wishful, fills a very important purpose in the learning algorithm that makes us intelligent.

Remember that temporal credit assignment is a challenge faced by all agents that need to learn from delayed rewards. The problem is essentially to figure out what actions leading up to a reward or penalty were responsible for the outcome and reinforce or weaken similar behavior in the future. In most work on RL, this problem is dealt with by relying on a temporal proximity heuristic where the actions closer in time to the reward state are straightforwardly assigned more credit than those farther away, i.e., the credit is somehow discounted based on the temporal proximity to the reward. A notable example of this is the board game playing system AlphaGo and its descendants (Silver et al., 2017; Silver et al., 2018) that use a hardcoded model of the world (the rules of the game) for planning and, when doing credit assignment, simply assign the same credit to all moves of a game leading up to a win or loss. This works in board games like Go or chess which have little causal invariance (most actions/turns influence the end state of the game), and where it is easy to generate a lot of independent action trajectories (e.g., chess games), but fails when actions have very different importance, and you cannot afford to fail multiple times, or in multi-task settings where it is not obvious which action influenced which reward state. For instance, imagine a Wall Street investor who assigns credit to his daily choice of a tie when he reaches the office in the morning and finds out that the value of one of his assets has crashed overnight. Superstition can in many cases be viewed as a failure of credit assignment.

Instead of relying only on temporal proximity, agents become more efficient learners if they are capable of figuring out what actions were actually important for a received reward or penalty. They must understand which actions had a causal influence on the reward. To this end, agents need to be able to imagine choosing other actions than those that were actually chosen. It comes down to answering counterfactual questions, such as ``What would have happened if I had not done X? Would Y still happen?" Accordingly, to figure out what actions caused a received reward or penalty, agents need to imagine alternative realities. Along with planning by simulating the future, the simulation of such alternative histories is a key benefit of having a world model that encodes causal relationships between variables. So-called hindsight credit assignment has been investigated (Harutyunyan et al., 2019, Mesnard et al., 2020; Guez et al., 2020), albeit, to the best of our knowledge, only for model-free RL, i.e. RL when there is no explicit world model. 

One can argue that not only credit assignment but also planning involves imagining taking actions independently of causal influence from the environment. When simulating future scenarios, agents need to evaluate the consequence of different decisions and can imagine being free to do virtually anything as long as it can be represented by their world model. For instance, Deery (2014) argued that prospection, i.e., our experience of different future scenarios through mental simulation, is the cause for people’s free will belief. However, since the action of planning lies in the causal path to the future, it is less clear that the imagination of being free in the future contradicts a deterministic model of the world and the agent itself. For instance, you can easily imagine doing something obviously stupid with a very low action value tomorrow, and hence imagine doing something that goes against your deterministic decision process. But if you actually go ahead and do this thing tomorrow, it is because reading this sentence and then thinking about what to do have skewed your value estimates to make it seem as if that thing is worth doing, and thus you have not escaped determinism. 

Since planning involves modeling the future, at least in its basic form, it does not require you to think about yourself as non-deterministic and having the ability to do differently than you actually did, i.e., no counterfactuals are needed. Although, it is not uncommon for humans to figure out what to do by imagining alternative versions of the present. One way to evaluate whether it is a good idea for us as a society to build an artificial general intelligence is to imagine an alternate reality where this has already been done. 

In summary, we argue that, in order to efficiently learn from our successes or mistakes, we need to imagine that we are free. This \textit{imagined freedom} is physical freedom in the simulated reality of our world model. When planning and performing credit assignment by counterfactual reasoning, we break the causal structure of our virtual reality and basically decide that something that did not in fact happen actually happened. In other words, we decide that events should occur without causes. The real causes for events in our mental simulation lie outside the simulation. In the frame of reference of the simulation, they are, to use physics jargon, hidden variables. Therefore, according to our imagination, we have physical freedom. Our ability to mentally break the chains of determinism makes us believe that—or at least feel as if—we really have physical freedom and that we can make decisions that are not ultimately caused by events external to ourselves.

\section{Discussion}

In this paper, we argued that centuries-long philosophical debate about free will is a result of a necessary, but flawed aspect of our self-concept. Our models of ourselves say that, when offered a choice, we could have chosen to do something that we did not in fact do. This model has a good purpose. Without it, we would be unable to apply counterfactual reasoning to the problem of credit assignment and efficiently learn from our successes and failures. To learn in complex situations, we need to imagine that we could have made other choices than those we actually made. And, interestingly, this model of ourselves is not something we seem free to get rid of. On some level, even the most determined determinists act as if they have free will. While they might intellectually know that they have no physical freedom, if they truly thought as if they did not, they would fail to learn efficiently.

Furthermore, we argue that this contradiction can be resolved by applying a reward maximization perspective on human decision-making. It allows us to formalize what \textit{free} and \textit{will }actually refer to and realize that we do not really need to believe that we have physical freedom. Instead, what we refer to here as value freedom suffices. When we reason about our own freedom, write laws, or communicate with each other in general, this is the kind of freedom we are actually referring to. Therefore, our freedom is not based on our ability to act independently from the world—rather, it emerges from our brain’s reward systems and the predictions of our world models. And since freedom tells us something about how much room for exploration and learning we have, it becomes, in Dennett’s (1984) parlance, ``worth wanting"—for humans, and for learning agents in general.

Of note, our definition of value freedom does not rely on RL being a complete and accurate theory of human learning and decision-making. There is no requirement for any learning to occur, and rewards do not necessarily need to be scalar. The only requirement is that agents should be capable of selecting actions based on their expected relative benefit. Such a selection process appears to be a condition for agency itself. To choose is to weigh different options against each other, and then make a decision. While the nature of this weighting can considerably vary, agency postulates the need for some kind of ordering of alternatives. Similarly, it is clear that humans are capable of performing complex credit assignments using some sort of causal model of the world.

In some sense, by clearly defining what freedom is, regardless of whether or not determinism holds, our conceptualization of value freedom allows for a compatibilist view on free will. We are not disagreeing with the \textit{classic incompatibilist argument (}McKenna $\&$ Coates, 2021) that determinism does not allow for free will; rather, what we propose is that the kind of freedom lay people actually refer to in everyday language is not the same kind of freedom incompatibilists reject. This distinction between what we here call physical freedom vs. value freedom allows for several potentially interesting perspectives.

For instance, when the UN declares that freedom of opinion is a human right, this does not imply that humans shall be able to choose their opinions independently from their environment. Rather, this means that people shall be able to do so without risking being punished for ``wrong" choices, or bribed into making ``correct" ones. When people vote to select their leaders, they should not perceive huge differences in estimated rewards between candidates. Although some differences, seemingly more or less important, must obviously exist, voters should know that making one choice or the other will not massively change their outlook. They will not get jail time, nor will they get a promotion, regardless of how they cast their votes. Said differently, freedom of opinion means that, when forming opinions, agents should not experience any extremes in the action value distribution.

From this perspective, political liberty should be seen not only as freedom from oppression, but also as freedom from excessively strong positive incentives. Since freedom stems from reward prediction, it has a dual nature where it can be restricted from both a negative and a positive direction. This duality comes in many well-known pairs, such as whips or carrots, fines or bribes, rape or prostitution. That we frequently consider only the negative part of this spectrum as potential restrictions of freedom is presumably due to the asymmetry between suffering and pleasure—an asymmetry that, on a fundamental level, likely arises from the irreversibility of death. Beyond suffering lies the risk of death, while pleasure has no such ultimate consequence. Beyond pleasure lies only more or less pleasure. It is very difficult to imagine an experience equally positive as the negative experience of slowly dying of physical torture. Therefore, to reduce freedom to the same extent as moderate penalties do, a reward must be very large.

Furthermore, the RL perspective on human decision-making also suggests that laws, be they written or moral, do not exist because they reflect some metaphysical justice or ideal, but rather because they are ways for societies of learning agents to teach each other how to act. Nothing is morally good or bad in itself, since the only value judgments that exist are action value estimates made by RL agents. Instead, the reason for societies to have rules is to shape the action value estimates of their members. Accordingly, the classic incompatibilist position—the one that submits that, without free will, no one can be morally responsible for their actions (McKenna $\&$ Coates, 2021)—is based on a misunderstanding of the purpose of assigning responsibility. Moral responsibility must be forward-looking (McKenna $\&$ Coates, 2021). The main reason for punishing members of a society, or any other group, for something they have done, is to discourage similar behavior in the future, and it makes little difference whether they were physically free to do what they did or not. From an RL perspective, what is important is to reduce the likelihood of someone making the same choice again in a similar situation by skewing future action value estimates. The ``responsibility assignment" that societies have created courts for can be viewed as analogous to the credit assignment single RL agents need to deal with when learning what behaviors to suppress and what behaviors to reinforce. In this view, courts are important components of the multi-agent learning system we call society.

There has been some empirical research on the connection between counterfactual thinking and free will belief. For instance, in a series of experiments, Alquist et al. (2015) found the belief in free will to be linked to increased counterfactual thinking. The authors then hypothesized that ``individuals and society as a whole may have benefitted from free will beliefs and the counterfactual simulations they stimulate". Here we show why this is the case and demonstrate not only that individuals and societies benefit from counterfactual thinking, but also how and why such thinking is essential for learning on a fundamental level through its role in efficient credit assignment.

Another observation is the relation between freedom and indifference, and how the latter is central to certain religious beliefs. In line with empirical accounts of laypeople's intuition about freedom (Deutschländer et al., 2017), an indifferent agent has high value freedom because it assigns equal value to multiple available options. However, indifference also means that the estimated action values are close to zero, while high freedom only implies that several options have approximately equal value. If an agent assigns a very high value to multiple options (but not all), that agent cannot be said to be indifferent, as it still thinks selecting one of these options is important. This relation between freedom and indifference has interesting philosophical analogies. According to the definition of value freedom, a maximally free agent is the one that always estimates all actions to have equal value. Such an agent never wants anything more or less than anything else, and obviously, would never learn anything and be utterly useless. One can question if it would be an agent at all. Applied to humans, such an agent is a person who completely lacks ego—an individual without any needs, drives, or desires. Interestingly, this view that the destruction of the ego is the ultimate path to freedom closely resembles a common theme in Eastern philosophy. According to Buddha’s \textit{Four Noble Truths}, to end personal suffering, stop the cycle of rebirth, and reach spiritual liberation, \textit{nirvana}, we must let go of all desires. It is only when we have no desires—not even the desire to rid ourselves of all desires—that we become completely free. Accordingly, nirvana is described as a state of complete emptiness. It is when we experience ultimate indifference, and end experience itself. Nirvana is thus a state of eternal maximum value freedom.

A venue for future research is to empirically investigate whether information entropy of the action selection distribution is an accurate measure of how people estimate freedom in simple decision scenarios. One can assume that it is more complex than that, and a considerable inter-individual variation can be expected. However, if the hypothesis proposed in this paper is correct, people’s freedom estimates should strongly correlate with the uniformity of their action value estimates.

\section{References}

Alquist, J. L., Ainsworth, S. E., Baumeister, R. F., Daly, M., $\&$ Stillman, T. F. (2015). The Making of Might-Have-Beens: Effects of Free Will Belief on Counterfactual Thinking. \textit{Personality and Social Psychology Bulletin}, \textit{41}(2), 268–283. doi:10.1177/0146167214563673

Coggan, M. (2004), ‘Exploration and exploitation in reinforcement learning’, Research supervised by Prof. Doina Precup, CRA-W DMP Project at McGill University.

Deepak Pathak, Pulkit Agrawal, Alexei A. Efros and Trevor Darrell. (2017) Curiosity-driven Exploration by Self-supervised Prediction. In ICML 2017.

Deery, O. (2014). Why people believe in indeterminist free will. Philosophical Studies, 172(8), 2033–2054. doi:10.1007/s11098-014-0396-7 

Deutschländer R, Pauen M, Haynes JD. (2017), Probing folk-psychology: Do Libet-style experiments reflect folk intuitions about free action? Conscious Cogn. 2017 Feb;48:232-245. doi: 10.1016/j.concog.2016.11.004. Epub 2016 Dec 23. PMID: 28013177.

Deisenroth, M. and Rasmussen, C. E. (2011). PILCO: A model-based and data-efficient approach to policy search. In Proceedings of the 28th International Conference on machine learning (ICML-11), pages 465–472.

Dennett, D. C. (1984). Elbow room : the varieties of free will worth wanting. Cambridge, Mass.: MIT Press. ISBN 0-262-04077-8.

Dennett, D. C. (2017), Reflections on Sam Harris' "Free Will". Rivista internazionale di Filosofia e Psicologia, [S.l.], v. 8, n. 3, p. 214-230, dic. 2017. ISSN 2239-2629.

Guez, A., Fabio Viola, Théophane Weber, Lars Buesing, Steven Kapturowski, Doina Precup, David Silver, $\&$ Nicolas Heess. (2020). Value-driven Hindsight Modelling.

Harutyunyan, A., Dabney, W., Mesnard, T., Gheshlaghi Azar, M., Piot, B., Heess, N., Hasselt, H., Wayne, G., Singh, S., Precup, D., $\&$ Munos, R. (2019). Hindsight Credit Assignment. In Advances in Neural Information Processing Systems.

Harris, S. (2012). Free will. Free Press.

Lai, T. and Robbins, H. (1985). Asymptotically efficient adaptive allocation rules. Adv. Appl. Math., 6(1):4–22. 14

Levine, S. and Koltun, V. (2013). Guided policy search. In International Conference on Machine Learning, pages 1–9.

McKenna, M., Coates, D.J. (2021), "Compatibilism", The Stanford Encyclopedia of Philosophy (Fall 2021 Edition), Edward N. Zalta (ed.)

Mesnard, T., Théophane Weber, Fabio Viola, Shantanu Thakoor, Alaa Saade, Anna Harutyunyan, Will Dabney, Tom Stepleton, Nicolas Heess, Arthur Guez, Marcus Hutter, Lars Buesing, $\&$ Rémi Munos. (2020). Counterfactual Credit Assignment in Model-Free Reinforcement Learning.

Minsky, M. L. (1963). Steps toward artificial intelligence. In E. A. Feigenbaum $\&$ J. Feldman (Eds.), Computers And Thought (pp. 406-450). New York, NY: McGraw-Hill.

Pearl J. (2010). An introduction to causal inference. The international journal of biostatistics, 6(2), 7. https://doi.org/10.2202/1557-4679.1203

Pereboom, D. (2001). Living without Free Will. Cambridge: Cambridge University Press. ISBN 9780521029964.

Schmidhuber, J.  (1991). A possibility for implementing curiosity and boredom in model-building neural controllers. In J. A. Meyer and S. W. Wilson, editors, \textit{Proc. of the International Conference on Simulation of Adaptive Behavior: From Animals to Animats}, pages 222-227. MIT Press/Bradford Books

Silver, D., Schrittwieser, J., Simonyan, K., Antonoglou, I., Huang, A., Guez, A., Hubert, T., Baker, L., Lai, M., Bolton, A., et al. (2017a). Mastering the game of go without human knowledge. Nature, 550(7676):354.

Silver, D., Hubert, T., Schrittwieser, J., Antonoglou, I., Lai, M., Guez, A., $\ldots$ Hassabis, D. (2018). A general reinforcement learning algorithm that masters chess, shogi, and Go through self-play. \textit{Science}, \textit{362}(6419), 1140–1144. doi:10.1126/science.aar6404

Silver, D., Singh, S., Precup, D., Sutton, R.S. (2021), Reward is enough, Artificial Intelligence, Volume 299, 2021, 103535, ISSN 0004-3702. doi:10.1016/j.artint.2021.103535

Thompson, W. R. (1933a). On the likelihood that one unknown probability exceeds another in view of the evidence of two samples. Biometrika, 25(3/4):285–294. 14

Wisniewski D, Deutschländer R, Haynes JD (2019) Free will beliefs are better predicted by dualism than determinism beliefs across different cultures. PLOS ONE 14(9): e0221617. 

United Nations (1948) Universal declaration of human rights.

\end{document}